\documentclass[review]{elsarticle}
\usepackage{xcolor}
\usepackage{lineno,hyperref}
\usepackage{graphicx}
\usepackage{subcaption}
\usepackage{algorithm}
\usepackage{algorithmic}
\usepackage{amsmath}
\modulolinenumbers[5]
\floatname{algorithm}{Procedure}

\journal{Journal of \LaTeX\ Templates}

\begin{document}

\begin{frontmatter}

\title{Deep Feature Selection using a Teacher-Student Network}


\author{Ali Mirzaei}
\ead{ali\_mirzaei@aut.ac.ir}

\author[]{Vahid Pourahmadi\corref{sdd}}
\ead{v.pourahmadi@aut.ac.ir}
\cortext[sdd]{Corresponding author}

\author{Mehran Soltani}
\ead{mehran73@aut.ac.ir}

\author{Hamid Sheikhzadeh}
\ead{hsheikh@aut.ac.ir}


\address{Amirkabir University of Technology, Tehran, Iran}




\begin{abstract}
High-dimensional data in many machine learning applications leads to computational and analytical complexities. Feature selection provides an effective way for solving these problems by removing irrelevant and redundant features, thus reducing model complexity and improving accuracy and generalization capability of the model. In this paper, we present a novel  teacher-student feature selection (TSFS) method in which a 'teacher' (a deep neural network or a complicated dimension reduction method) is first employed to learn the best representation of data in low dimension. Then a 'student' network (a simple neural network) is used to perform feature selection by minimizing the reconstruction error of low dimensional representation. Although the teacher-student scheme is not new, to the best of our knowledge, it is the first time that this scheme is employed for feature selection. The proposed TSFS can be used for both supervised and unsupervised feature selection. This method is evaluated on different datasets and is compared with state-of-the-art existing feature selection methods. The results show that TSFS performs better in terms of classification and clustering accuracies and reconstruction error. Moreover, experimental evaluations demonstrate a low degree of sensitivity to parameter selection in the proposed method.
\end{abstract}


\end{frontmatter}

\section{Introduction}

The dimensionality of captured data in common applications is increasing constantly, as for example, an image taken from a typical camera has nearly one million features. Although it is prohibitively expensive to directly process high-dimensional data, often most of the features are redundant or highly correlated. Accordingly, the “intrinsic dimensionality” of data is often much lower than the original feature space. The dimension reduction has several advantages: 1) Data storage is reduced, 2) Machine learning models are simplified which leads to increased generalization capability, 3) Computational complexity for training and testing of machine learning models are reduced. 

Dimension reduction techniques are generally divided into two categories: feature extraction and feature selection. Feature extraction methods try to find a linear or non-linear projection which maps the original high-dimensional data into a lower-dimensional subspace. These methods employ \emph{all} features to obtain an optimal representation of data which does not necessarily have any semantic and is usually hard to interpret. On the other hand, feature selection methods aim to select a subset of the original high-dimensional features based on some performance criterion. Therefore, they can preserve the semantics of  the original features and produce dimensionally reduced results that are more interpretable for domain experts. Another advantage of feature selection is that once the low-dimensional features have been selected, one only needs to collect or calculate the selected features during data acquisition. This can be helpful because sometimes the measurement of all feature are costly or practically impossible.
 
Based on \emph{training data} (labeled or unlabeled), feature selection methods can be classified into
three categories: supervised \cite{zhang2018self, gan2018supervised} , semi-supervised \cite{sechidis2018simple, sheikhpour2017survey} and unsupervised \cite{panday2018feature, li2018unsupervised, qi2018unsupervised, zhu2018co}. Moreover, based on their \emph{relationship with learning models}, they can be divided into three categories: wrapper, filter and embedded methods \cite{mirzaei2017variational, mohsenzadeh2016incremental}. In wrapper and embedded methods, feature selection is tied to learning models. The difference is that in wrapper methods all subset of features are evaluated on model and the best subset of features is selected based on model performance. In contrast, in embedded methods model training and feature selection are performed simultaneously. Finally, filter methods try to rank the features based on their importance according to a specific criteria. Usually filter methods have much less computational complexity compared to wrapper and embedded approches.   

Recently deep-learning has gained much attention, so that in most applications like computer vision, natural language processing (NLP) and audio processing, the state-of-the-art algorithms are based on deep-learning. In a deep-learning approach, a deep neural network extracts the best representation of data for a specific task. This provides us with motivation to investigate the application of deep models for feature selection. 

There are some studies that try to use neural network for feature selection \cite{han2018autoencoder, feng2018graph, chang2017dropout}. The AEFS method \cite{han2018autoencoder} uses a single-layer autoencoder to reconstruct data and perform feature selection with row-sparsity constraint on the first layer of autoencoder. In GAFS \cite{feng2018graph}, similar to AEFS, a single layer autoencoder is used for data reconstruction and feature selection. Additionally, they incorporate the spectral graph analysis of the projected data into the learning process. This technique preserves the local data geometry of the original data space in the low-dimensional feature space.

All mentioned neural network based feature selection methods employ a simple auto-encoder to perform feature selection based on reconstruction error. Actually, for feature selection it is necessary to assume a simple network structure to make sure the error can be easily back-propagated and the best features can be learned effectively. The important point is that this simple network is also used for reconstruction of original data. Such simple network can not model the complex manifold of data and leads to high reconstruction error and non-optimal feature selection (complex non-linear correlated features are selected together).

\begin{figure}
    \centering
    \includegraphics[width=\textwidth]{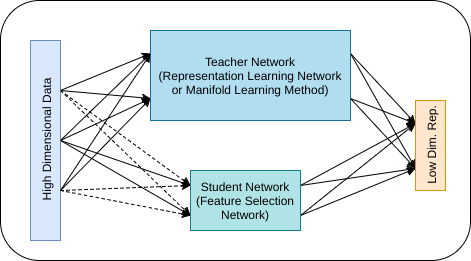}
    \caption{The proposed teacher-student scheme for feature selection}
    \label{fig:teacher-student}
\end{figure}

Motivated to solve this problem, in this paper we propose a teacher-student scheme. In essence, we face with two problems: feature selection and data reconstruction. For feature selection it is necessary to have a simple network so that error can be easily back-propagated and best features can be learned effectively. On the other hand, data reconstruction needs a complicated network to learn the complex manifold of data and reconstruct data using limited selected features. To address both issues simultaneously, we employ a teacher-student structure. The teacher, performing a \emph{feature extraction}, uses whatever complex method it desires, to learn a good low-dimensional representation of data. After that, the student network (which has a simple structure) is no longer concerned with finding good low-order representation and instead tries to mimic the output of the teacher and performs \emph{feature selection}. Figure \ref{fig:teacher-student} shows a schematic of the proposed approach. The teacher-student scheme of  \cite{hinton2015distilling} previously was used to train a small and fast network for classification problems and in this study we employ it for feature selection. The proposed method, Teacher Student Feature Selection (TSFS), can be used for both supervised and unsupervised feature selection (based on teacher method). In this paper, we extensively evaluate TSFS on different datasets and the performance results are compared with state-of-the-art existing feature selection methods. The results show the TSFS performs better in terms of classification and clustering accuracies and reconstruction error. Moreover, the sensitivity of the algorithm with respect to its parameters is  experimentally evaluated and it is shown that the method has a low parameter sensitivity.


The rest of this paper is organized as follow: Section \ref{section:related} explains the related works on most types of feature selection methods and also describes the knowledge distillation concept which is the base of this research. In Section \ref{method} the proposed method is detailed. Section \ref{section:evaluation} evaluates the proposed method and compares it with existing methods and finally Section \ref{section:conclusion} concludes the paper. 

\section{Notations and Related Works}
\label{section:related}

In this section, we introduce needed notations used in this paper and provide a review of major related works. At the end of this section knowledge distillation concept in deep neural networks are described.
\subsection{Notations}
In this paper bold capital letters represent matrices, small bold letters are vectors and normal letters are scalar values.
The datasets are denoted by $\mathbf{X} \in R_{n\times d}$ , where $\mathbf{x}_i \in R_d$ is the $i_{th}$ sample in $\mathbf{X}$ and $d$ denotes number of features (number of columns of $\mathbf{X}$) and $n$ denotes the number of samples (number of rows of $\mathbf{X}$). Also $x_{ij}$ is the value of the $i_{th}$ row and $j_{th}$ column. 

The p,q-norm for a matrix $\mathbf{W} \in R_{n\times m}$ is defined as:

\begin{equation}
\label{eq:norm}
     ||\mathbf{W}||_{p,q} = (\sum_{i=1}^n(\sum_{j=1}^m|w_{ij}|^p)^{\frac{q}{p}})^{\frac{1}{q}}.
\end{equation}

The most useful norms are $||\mathbf{W}||_{2,2}$ and $||\mathbf{W}||_{2,1}$. The $||\mathbf{W}||_{2,2}$ is known as Frobenius norm. $||\mathbf{W}||_{2,1}$ applies the $l_2$ norm on column elements and $l_1$ norm on the calculated norm of rows. When this norm is used as a regularizer it leads to sparsity in rows of $\mathbf{W}$.

\subsection{Feature Selection Methods}
\label{subsection:fsmethods}
Over the past three decades, many feature selection algorithms have been proposed. These methods can be generally grouped to four main categories: similarity based, information theoretical based, statistical based and sparse learning based  methods. A brief description of each category follows here. 
\subsubsection{Similarity Based Feature Selection}
\label{subsection:similaritybasedmethods}
Similarity based methods evaluate feature
importance by their ability to preserve data similarity. The Laplacian Score (LS)  \cite{he2006laplacian} is a well-know unsupervised similarity-based feature selection. The basic idea of LS is to evaluate the features according to their locality preserving power. This method constructs a nearest neighbor graph for dataset samples and use the Laplacian matrix to calculate a score for each feature which shows the importance of the feature in preserving the similarity and locality of dataset samples.

These methods are straightforward and simple to implement and also they are independent of the learning approach and thus the selected features can be used for any subsequent learning tasks. However, one drawback of these methods is that they cannot handle feature redundancy. In other words, they may select a highly correlated feature set during the selection phase. 

\subsubsection{Information Theoretical based Feature Selection}

Unlike similarity based feature selection algorithms that are unable to handle feature redundancy, the information theoretical based feature selection algorithms use the information theory to consider both \emph{feature relevance} and \emph{feature redundancy}. Most of these methods, use the label data to consider the relevance and redundancy of features and so these feature are usually supervised.
One the well-known representative of this family is Minimum Redundancy Maximum Relevance (MRMR) \cite{peng2005feature}. This method selects an optimal feature set according to the maximal statistical dependency criterion based on mutual information.

Similar to methods of Section \ref{subsection:similaritybasedmethods}, this category is independent of subsequent learning algorithms. However, most of the existing information theoretical based feature selection methods can only be used in a supervised scenario and label availability is necessary.

\subsubsection{Statistical based Feature Selection}
Algorithms in this category of feature selection methods are based on various statistical measures. Since in these methods the features are usually evaluated individually, the correlation of the features can not be captured. Variance-score, t-score and chi-score are well-known representatives of this family.

The complexity and computational load of these methods are often very low. Therefore, they are often used as a pre-processing step before applying other advanced feature selection algorithms.

\subsubsection{Sparse Learning based Feature Selection}
An interesting and practical class of feature selection method is the one based on sparse learning. These methods aim to preserve subspace structure as much as possible. The underlying subspace structure of data are usually different in various applications. The basic idea is to use a transformation matrix to project data to a new space and perform feature selection based on the sparsity of the transformation matrix employing an objective function to be optimized. The sparsity can be realized with adding a regularizer to the objective function. This sparse regularizer forces many feature coefficients to be small, or exactly zero, and then the corresponding features (with small or zero values) can be simply eliminated. Sparse learning based methods have received much attention in recent years due to their good performance and interpretability \cite{cai2010unsupervised}, \cite{zhu2015unsupervised}, \cite{han2018autoencoder}, \cite{feng2018graph}.
In the following we present major new and related works.

\begin{itemize}
    \item MCFS \cite{cai2010unsupervised}: Multi-clustering feature selection (MCFS) creates a $k$-nearest weighted graph for all samples. Then it uses the spectral graph analysis and map the vertices (samples) to a low-dimensional subspace. This method is based on the following optimization term:
    
    \begin{equation}
    \label{eq:mcfs}
        \min ||\mathbf{Y - WX}||^2_F + \lambda||\mathbf{W}||_{1,1},
    \end{equation}
where $\mathbf{Y}$ is a low-dimensional subspace obtained by spectral graph analysis, F-norm is the Frobenius norm and $||W||_{1,1}$ is the norm defined by Eq. \eqref{eq:norm}.

One of the main disadvantageous of this method is that it uses a linear model (as evident in \eqref{eq:mcfs}) to approximate the subspace samples using original features.

    \item RSR \cite{zhu2015unsupervised}: In this paper, the authors claim that it is difficult to choose the proper subspace matrix ($Y$) in \eqref{eq:mcfs}. They propose a regularized self-representation (RSR) model for unsupervised feature selection. In other words, this method simply replaces the data matrix $X$ for the subspace matrix $(Y)$. It means a feature is calculated based on a linear combination of other features and the features contributing more to construction are retained.
    
    In this method the objective function is defined as:
    
    \begin{equation}
        \label{eq:rsr}
        \min ||\mathbf{X - WX}||_{2,1} + \lambda||\mathbf{W}||_{2,1},
    \end{equation}
    
    for both loss function and the regularizer parts of Eq. \eqref{eq:rsr}, the $L_{2,1}$ norm  is used (is defined in Eq. \eqref{eq:norm}).
The $L_{2,1}$ regularizer consequently leads to row-sparsity of weight matrix. Finally, the score of each feature is computed based the Euclidean norm of its corresponding row in the weight matrix. 

Similar to MCFS, this method uses a linear transformation to reconstruct data. 
  
    \item AEFS \cite{han2018autoencoder}: The AutoEncoder inspired unsupervised Feature Selection (AEFS) is a more general version of RSR method where it assumes a non-linear relation between features. It uses a single-layer autoencoder to reconstruct data. In this method the optimization problem is defined as:
    
    \begin{equation}
        \label{eq:aefs}
        \min ||\mathbf{X} - g(f(\mathbf{X}))||^2_F + \lambda||\mathbf{W^{(1)}}||_{2,1} + \beta\sum_{i=1}^2 \mathbf{W^{(i)}} ,
    \end{equation}
    
where $\mathbf{W^{(1)}, W^{(2)}}$  are the weight matrices of first and second layer of autoencoder, $f$ is the encoder function defined as $f(\mathbf{X})=\sigma_1(\mathbf{X}\mathbf{W}^{(1)})$ and $g$ is the decoder function that produces a reconstruction $\hat{\mathbf{X}} = g(f (\mathbf{X})) = \sigma_2(f(\mathbf{X})\mathbf{W}^{(2)})$, and $\sigma_1$ , $\sigma_2$ are
activation functions (e.g. sigmoid, ReLU, tanh) of the hidden layer and the output layer, respectively. 

This approach is able to uncover the existing nonlinear relationships between features. One of the main drawback of this method stems from the simplicity of autoencoder where a simple single-layer autoencoder cannot model complex non-linear dependencies of features.  
    \item GAFS \cite{feng2018graph}: graph autoencoder-based unsupervised feature selection (GAFS), similar to AEFS, tries to select features based on the reconstruction of the original data using the single-layer autoencoder. Additionally, GAFS uses a spectral graph analysis on the projected data in order to make sure that local data geometry is preserved in the low-dimensional feature space. Here, the objective function is written as:
    \begin{equation}
        \label{eq:gafs}
        \min ||\mathbf{X} - g(f(\mathbf{X}))||^2_F + \lambda||\mathbf{W^{(1)}}||_{2,1} + \beta Tr(\mathbf{YLY^t}) ,
    \end{equation}
    where $\mathbf{Y}$ is the output of the hidden codes of the autoencoder and $\mathbf{L}$ is the Laplacian matrix of the constructed graph. $Tr(.)$ is the trace operator and $Tr(\mathbf{YLY^t})$ shows a local data geometry preservation criteria.
    
    Similar to AEFS, this method also suffers from the simplicity of the reconstruction network as the autoencoder cannot handle all data types. For example, for image data one needs a complex convolutional network to reconstruct data. 
\end{itemize}

\subsection{Knowledge Distillation}
\label{knowledge}
It is possible to \emph{distill} the large and complicated neural network knowledge into another much smaller network where the smaller network tries to mimic the original function of the complex network. 

In this scheme, there are two networks: \emph{teacher} and \emph{student} networks. The teacher network has a complex structure which uses the hard labels of data to obtain a low dimension (soft labels) for the original data. Usually soft labels have a higher entropy and provide much more information per training case than hard labels. As a result, compared to the original complex model, a small model can often be trained on much less data. After obtaining the soft labels, the student network tries to mimic the output of the teacher network.   

It is shown that the performance the \emph{student} is comparable with the \emph{teacher} \cite{hinton2015distilling}. Using this scheme, the student can learn the teacher knowledge without much effort resulting in simpler architecture and faster learning and testing phases.

\section{Proposed Teacher-Student Feature Selection}
\label{method}

\subsection{Concept}
As mentioned in the previous section, the existing methods like AEFS and GAFS uses a simple autoencoder for both reconstruction and feature selection. For feature selection it is necessary to have simple network so that error can be easily back-propagated and the important features in input layer be selected efficiently. On the other hand, a single-layer autoencoder is not sufficient to reconstruct all types of data. For example for reconstruction and generating images, convolutional neural networks are best choices \cite{radford2015unsupervised}.  

To address this contradictory requirements, we propose a novel two-stage feature selection approach. In the first stage, all training data is mapped to a low dimensional space, and in the second stage, feature selection is performed based on new subspace of data. \textcolor{black}{Since the two tasks are now separated, we can now employ complex and simple network structures for the first and the second stages, respectively}. This approach is inspired from \emph{knowledge distillation} concept introduced in Section \ref{knowledge}. 

The two stages of the proposed method are described as follows:

\begin{figure}
    \centering
    \includegraphics[width=.9\textwidth]{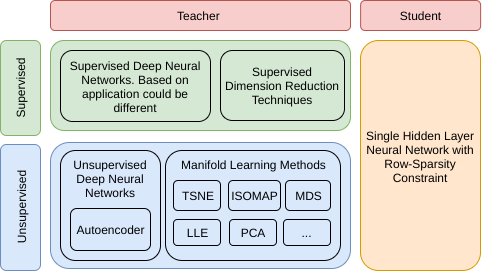}
    \caption{The teacher and the student approaches which could be select based on application for supervised and unsupervised tasks }
    \label{fig:TSFS}
\end{figure}

\textbf{Teacher Stage}: The teacher is usually a complex method which tries to obtain the best low dimensional representation of data. Normally, the teacher for different applications should be different based on the problem requirements and can be either supervised or unsupervised as shown in Figure \ref{fig:TSFS}. Fore example, a \emph{manifold-learning}  or \emph{deep-learning} method can be employed as the \emph{teacher} method. The teacher functionality is also shown by the top branch of Figure \ref{fig:teacher-student}.  

\textbf{Student Stage}: In the second stage, the low-dimensional codes are normalized and a \emph{student} network is trained based on these low-dimensional codes. We use a simple single fully-connected layer neural network for the student as shown in Figure \ref{fig:TSFS}. 

For feature selection, similar to the other works \cite{cai2010unsupervised,han2018autoencoder,feng2018graph}, we apply row-sparsity constraints on weights of the first layer of the network, as shown in bottom branch of Figure \ref{fig:teacher-student}. Finally, the features are ranked based on feature scores (described below) and features with highest scored are then selected.

\subsection{Algorithm}

Assume $\mathbf{X} \in R_{n\times d}$ is a matrix including the original dataset containing $n$ samples and $d$ features. We aim to select the top $p$ percent of features as most important features. To this end we perform three steps:

\textbf{1- Teacher Step}:
At the first step, a deep-learning network or a manifold learning approach is employed to obtain the best representation of data ($\mathbf{Y}$) in low dimensions. 
\begin{equation}
    \label{eq:teacher}
    \mathbf{Y}_{n\times l} = F(\mathbf{X}_{n \times d}),
\end{equation}
where $l$ is the dimension of latent space and usually $l \ll d$ and $F$ is a complicated and non-linear function. A dimension reduction technique, manifold learning technique or deep representation learning can be used for $F$ as shown in Figure \ref{fig:TSFS}. 

If $F$ is an \emph{unsupervised} method, then the proposed approach will be an unsupervised feature selection method. Alternatively, for a \emph{supervised} $F$, the proposed approach will be a supervised feature selection. After the $F$ is trained (in case of deep-learning approaches) or selected (in case of manifold approaches), all training data are fed to this function and the corresponding low-dimensional codes of data are obtained ($\mathbf{Y}$ matrix in Eq. \eqref{eq:teacher}).

\textbf{2- Student Step}:
In the feature selection stage, we train a single-layer neural network to reproduce the latent codes form limited selected features. For better training, the latent codes are normalized between 0 and 1.
\begin{equation}
    \mathbf{Y}_{n} = \frac{\mathbf{Y} - min(\mathbf{Y})}{max(\mathbf{Y})-min(\mathbf{Y})}  
\end{equation}

The output of the neural network can be written as:
\begin{equation}
    \mathbf{Y}_{p} = (Relu(\mathbf{XW^1}+\mathbf{b}^1))\mathbf{W^2}+\mathbf{b}^2,
\end{equation}
where $\mathbf{W^1}$ ($d\times h$) and $\mathbf{W^2}$ ($h\times l$) are the weight matrices of hidden and output layer of network, respectively. $\mathbf{b^1}$ and $\mathbf{b^2}$ are the bias vectors for network layers. 
To perform feature selection, we apply a row-sparsity constraint similar to \cite{han2018autoencoder,feng2018graph} as follows:

\begin{equation}
    ||\mathbf{W}^{1}||_{2,1} = \sum_i^d\sqrt{\sum_j^h(\mathbf{W}^{1}_{ij})^2}.
\end{equation}

Where $\mathbf{W}^{1}$ is the weight matrix of the first layer with $d$ rows and $h$ columns. The neural network loss function is defined as:

\begin{equation}
    J(\Theta)=\frac{1}{2n}||\mathbf{Y}_{n}-\mathbf{Y}_{p}||+\lambda||\mathbf{W}^1||_{2,1},
\end{equation}
where $\lambda$ is the trade-off parameter between the reconstruction loss and the regularization term.

\textbf{3- Feature Selection Step}:
After training of the student network, the importance scores of features are calculated as follows:

\begin{equation}
    \mathbf{s} =  diag(\mathbf{W}^1(\mathbf{W}^1)^T)
\end{equation}
where $\mathbf{s}$ is a $d$ dimensional vector containing the feature importance weights. To retain $p$ percent of features, the top $p\%$ of $\mathbf{s}$ will be selected.

The complete TSFS algorithm is illustrated in Procedure 1. The source code for TSFS is also avilable at \url{https://github.com/alimirzaei/TSFS}.

\begin{algorithm} \label{alg}
	\caption{Teacher Student Feature Selection Algorithm}
	\begin{algorithmic}
		\REQUIRE data: $\mathbf{X}_{n\times d}$, labels(optional): $\mathbf{T}_{n\times c}$ , percent of features $p \in (0,100) $
		\ENSURE selected features: $\{f_1, f_2, ..., f_m\}, f_i \in \{1, ..., d\}, where m=p*d/100 $
		\IF{teacher is \emph{deep-learning} based}
		    \STATE{Train the deep neural network using $\mathbf{X, T}$}
		    \STATE{Feed $\mathbf{X}$ to the network and obtain the learned representation($\mathbf{Y}$)}
		\ELSIF{teacher is \emph{dimension reduction} method}
		    \STATE{Feed $\mathbf{X}$ to the selected method and obtain the lower dimension($\mathbf{Y})$}
		\ENDIF
		
		\STATE{Train student network with $\mathbf{X,Y}$}
		\STATE{Calculate the feature scores based on first layer weights: $\mathbf{s}=diag(\mathbf{W^1(W^1)^T})$}
		\RETURN {The first $m$ values of $argsort_{descending}(\mathbf{s})$}

	\end{algorithmic}
\end{algorithm}

\subsection{Optimization}
For training the teacher and the student networks, we employed the Adam optimizer \cite{kingma2014adam} which can be used instead of the classical stochastic gradient descent (SGD) procedure to update network weights in an iterative manner. The method is based on adaptive estimates of lower-order statistical moments. The advantages of this algorithm are: 1) Ease of implementation, 2) computational and storage efficiency, 3) Fitness to the problems with large data or parameters \cite{kingma2014adam}.

\section{Evaluation}
\label{section:evaluation}
In this section we evaluate the TSFS on several real datasets from different aspects. Six various datasets are employed to show the performance of proposed algorithm for classification, clustering and reconstruction applications using four metrics. In first part of this section, the selected datasets are described, in second subsection the evaluation metrics are introduced and in third subsection, the performance of algorithm, compared with other existing methods, are illustrated. Finally, the last subsection performs a sensitivity analysis for the algorithm parameters.

\subsection{Dataset Description}
We evaluate the proposed method on different datasets \footnote{All datasets were downloaded from http://featureselection.asu.edu/datasets.php}. We use three text datasets: BASEHOCK, PCMAC and RELATHE, three image datasets: COIL20 and Yale and a MNIST subset and one audio dataset: Isolet. Detail specifications of datasets are shown in Table \ref{tab:datasets}. 

\begin{table}[]
    \centering
    \begin{tabular}{c|cccc}
         Dataset &  Instances &  Features & Classes & Type \\ \hline
    BASEHOCK & 1993&	4862&	2& Text \\ 
    PCMAC &	1943 & 3289 & 2 & Text \\ 
    RELATHE & 1427 & 4322 & 2 & Text \\ 
    COIL20 & 1440 & 1024 & 20 & Object Image \\ 
    Yale & 165 & 1024 & 15 & Face Image \\
    MNIST\_subset & 1000 & 784 & 10 & Handwritten Image \\
    Isolet & 1560& 617 & 26 & Audio \\
    \end{tabular}
    \caption{The detail specification of datasets used for evaluation of the TSFS}
    \label{tab:datasets}
\end{table}

Only for MNIST dataset, a subset containing 100 samples for each class (totally 1000 samples) are selected to remove the unbalanced number of samples in each class. Additionally, the computation load of some of comparing methods in our experiments were prohibitively heavy for the complete MNIST dataset.

\subsection{Evaluation Metrics}
\label{section:metrics}

In this study, we perform both supervised (classification) and unsupervised (clustering and reconstruction) evaluations on datasets using different percentage of selected features in order to evaluate the effectiveness of feature selection algorithms. For classification, we employ MLP classifier due to its simplicity and report the classification accuracy as the evaluation metric for feature selection effectiveness. Because some of the employed datasets are not splited to train and test subsets, we use 5-fold cross validation on original datasets and report the mean accuracy of the five test parts. 
\begin{table}[]
    \centering
    \begin{tabular}{c|c|c}
        Evaluation & Method & detail \& parameters \\ \hline
        Classification & MLP & 100 \emph{Relu} hidden neurons, Adam optimizer\\
        Clustering & K-Means & 20 Random initialization \\
        Reconstruction & MLP & 10 \emph{Relu} hidden neurons, Adam optimizer 
    \end{tabular}
    \caption{The evaluation methods for different percentage of selected features}
    \label{tab:evaluation}
\end{table}

For clustering, we use k-means clustering on the selected features. To evaluate clustering performance, two different evaluation metrics are used similar to previous studies of \cite{feng2018graph, han2018autoencoder}. The first is clustering accuracy (ACC), defined as: 

\begin{equation}
    ACC = \frac{1}{n} \sum_{i=1}^n\delta(y_i, map(c_i)),
\end{equation}
where $n$ is the total number of data samples, $\delta(a,b) = 1$ when $a = b$ and $0$ when $a \neq b$ , $map(.)$ is the optimal mapping function between cluster labels and class labels obtained using the Hungarian algorithm \cite{kuhn1955hungarian}, and $c_i$ and $y_i$ are the clustering and ground truth labels of a given data sample $\mathbf{x_i}$, respectively. The second metric is the normalized mutual information (NMI), defined as:

\begin{equation}
    NMI = \frac{MI(\mathbf{c}, \mathbf{y})}{max(H(\mathbf{c}), H(\mathbf{y}))},
\end{equation}
where $\mathbf{c}$ and $\mathbf{y}$ are clustering labels and ground truth labels, respectively, $MI( \mathbf{c}, \mathbf{y} )$ is the mutual information between $\mathbf{c}$ and $\mathbf{y}$ and $H ( \mathbf{c} )$ and $H ( \mathbf{y} )$ denote the entropies of $\mathbf{c}$ and $\mathbf{y}$, respectively. For both ACC and NMI, the clustering is performed for 20 times with random initialization for each case, and the corresponding mean values of ACC and NMI are reported.

Moreover, in order to show to what extent the selected features are chosen correctly, we try to reconstruct the original data using the selected features and report the mean square error (MSE). A single hidden layer neural network with $10$ neurons with \emph{Relu} activation function is employed. In all datasets, the number of input layer nodes are equal to selected features while the number of output layer nodes are equal to number of all features. 5-fold cross-validation is used to report the MSE. 

The detailed algorithms for evaluation are summarized in Table \ref{tab:evaluation}.

\begin{figure}
    \centering
    \includegraphics[width=\textwidth]{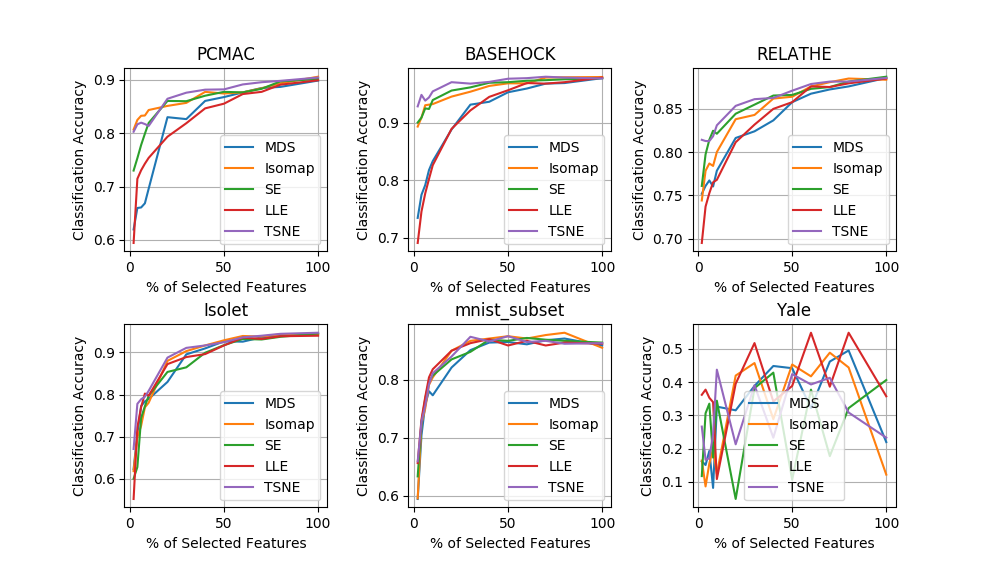}
    \caption{Classifcation accuracy using TSFS with manifold learning methods as the teacher, includes TSNE, Isomap, LLE, MDS and Specteral Embedding(SE) }
    \label{fig:my_results}
\end{figure}

\begin{figure*}[t!]
    \centering
    \begin{subfigure}[t]{0.33\textwidth}
        \centering
        \includegraphics[width=\textwidth]{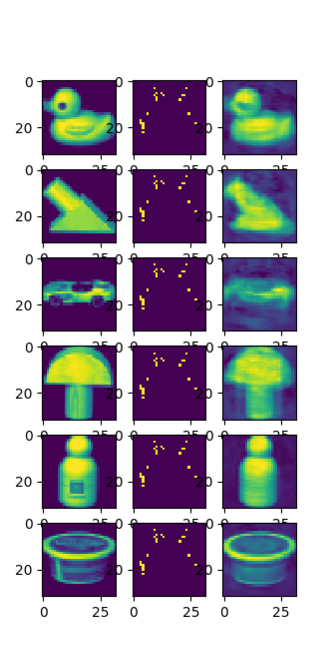}
        \caption{2\% of features}
    \end{subfigure}%
    \begin{subfigure}[t]{0.33\textwidth}
        \centering
        \includegraphics[width=\textwidth]{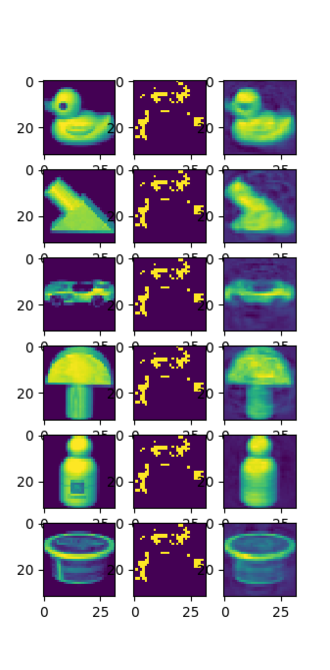}
        \caption{10\% of features}
    \end{subfigure}
    \begin{subfigure}[t]{0.33\textwidth}
        \centering
        \includegraphics[width=\textwidth]{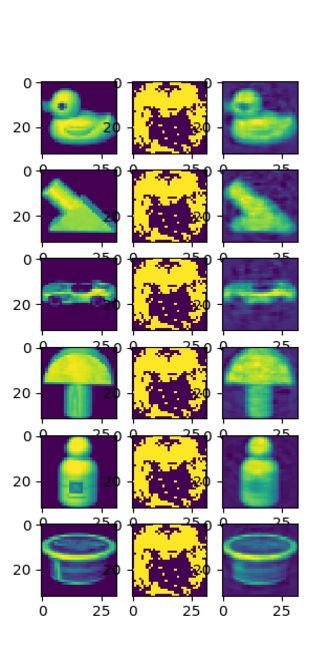}
        \caption{50\% of features}
    \end{subfigure}
    \caption{Image reconstruction using limited selected features. The first column is original image, the second column is selected features and the third column is reconstructed image using selected features}
    \label{fig:reconst}
\end{figure*}

\subsection{Performance Evaluation}
In this section we evaluate the TSFS performance for both unsupervised and supervised tasks. In first subsection the unsupervised version of our proposed method (U-TSFS) is compared with other existing methods in terms of classification, clustering and reconstruction error. The second subsection compares our supervised version of our proposed method (S-TSFS) with other methods in terms of classification accuracy.
\subsubsection{Unsupervised Teacher-Student Feature Selection(U-TSFS)}
\label{section:unsupervised}

\begin{figure}
    \centering
    \includegraphics[width=.9\textwidth]{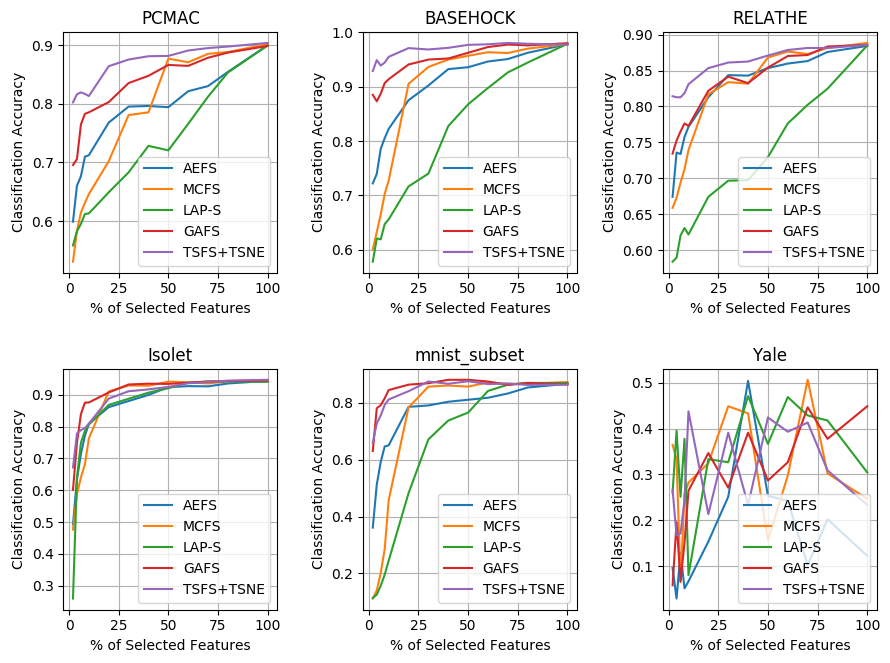}
    \caption{Comparison with other methods in terms of classification accuracy}
    \label{fig:res_classification}
\end{figure}

\begin{figure}
    \centering
    \includegraphics[width=.9\textwidth]{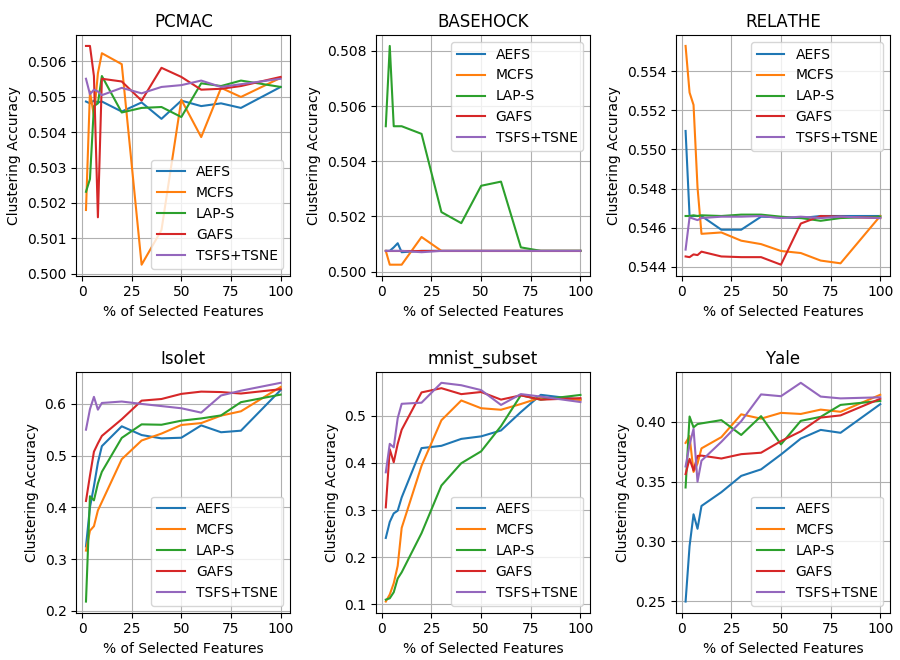}
    \caption{Comparison with other methods in terms of clustering accuracy}
    \label{fig:res_clustering}
\end{figure}

\begin{figure}
    \centering
    \includegraphics[width=.9\textwidth]{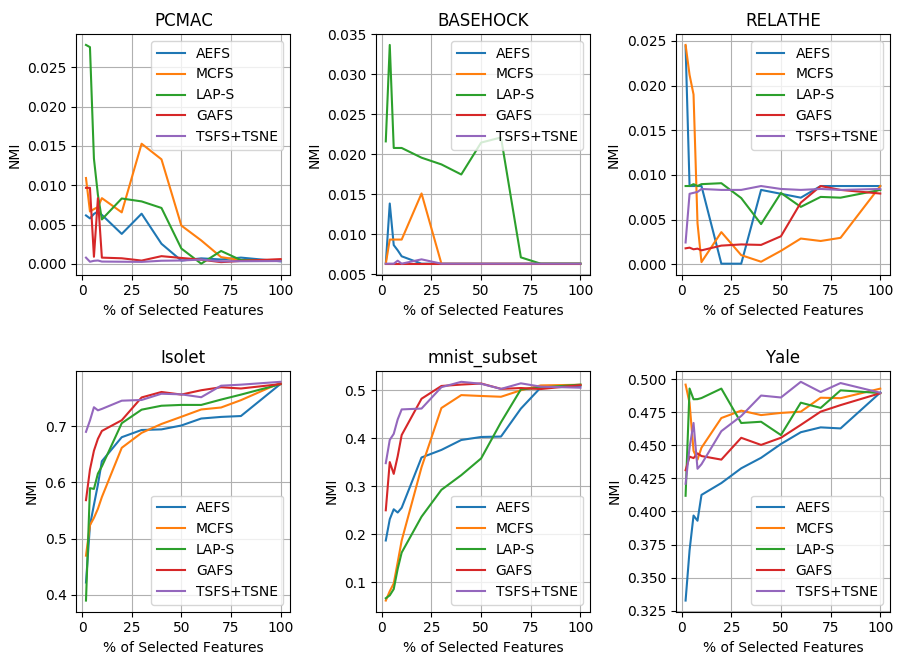}
    \caption{Comparison with other methods in terms of NMI}
    \label{fig:res_NMI}
\end{figure}

\begin{figure}
    \centering
    \includegraphics[width=.9\textwidth]{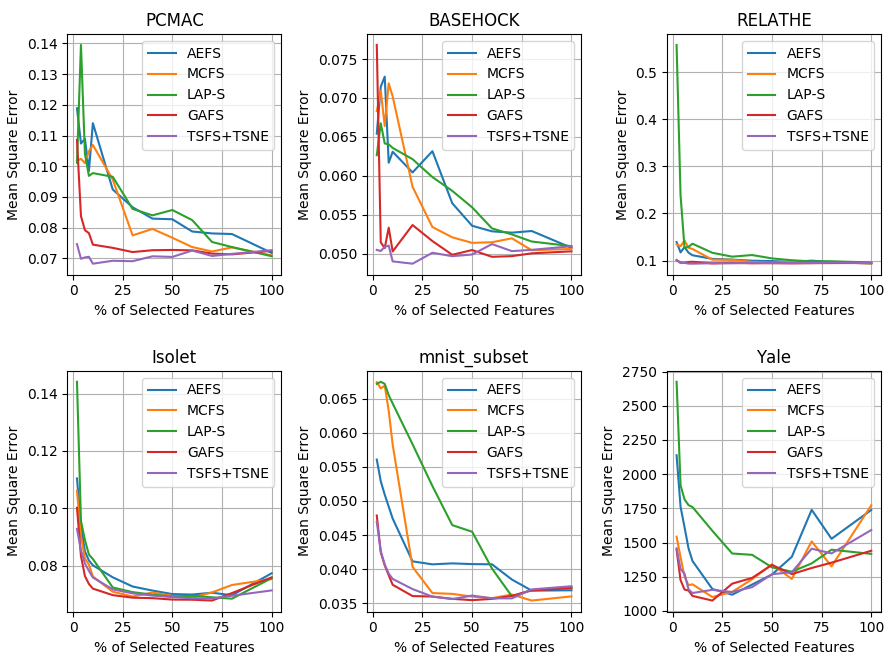}
    \caption{Comparison with other methods in terms of reconstruction error (MSE)}
    \label{fig:res_mse}
\end{figure}


\begin{figure}
    \centering
    \includegraphics[width=.7\textwidth]{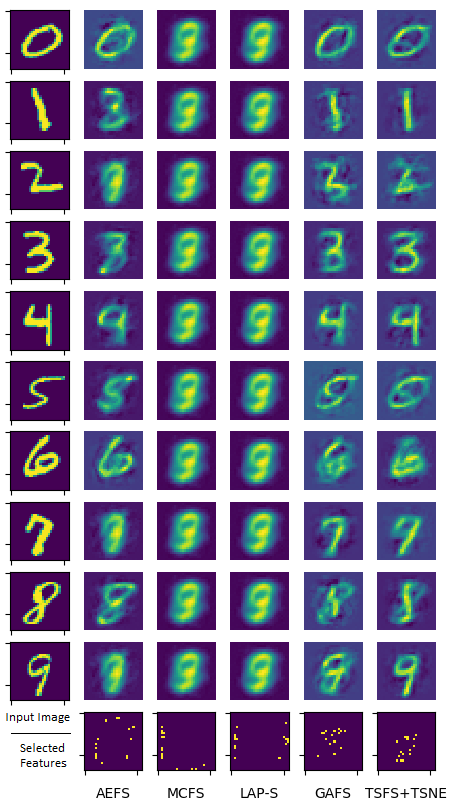}
    \caption{Visual reconstruction performance comparisons with other methods. 15 features are selected and the images are reconstructed using only these 15 features. The reconstruction method is the same for all feature selection approaches. The first column is the original image and the last row shows the selected feature for different feature selection approaches}
    \label{fig:mnist}
\end{figure}

\begin{figure}
    \centering
    \includegraphics[width=\textwidth]{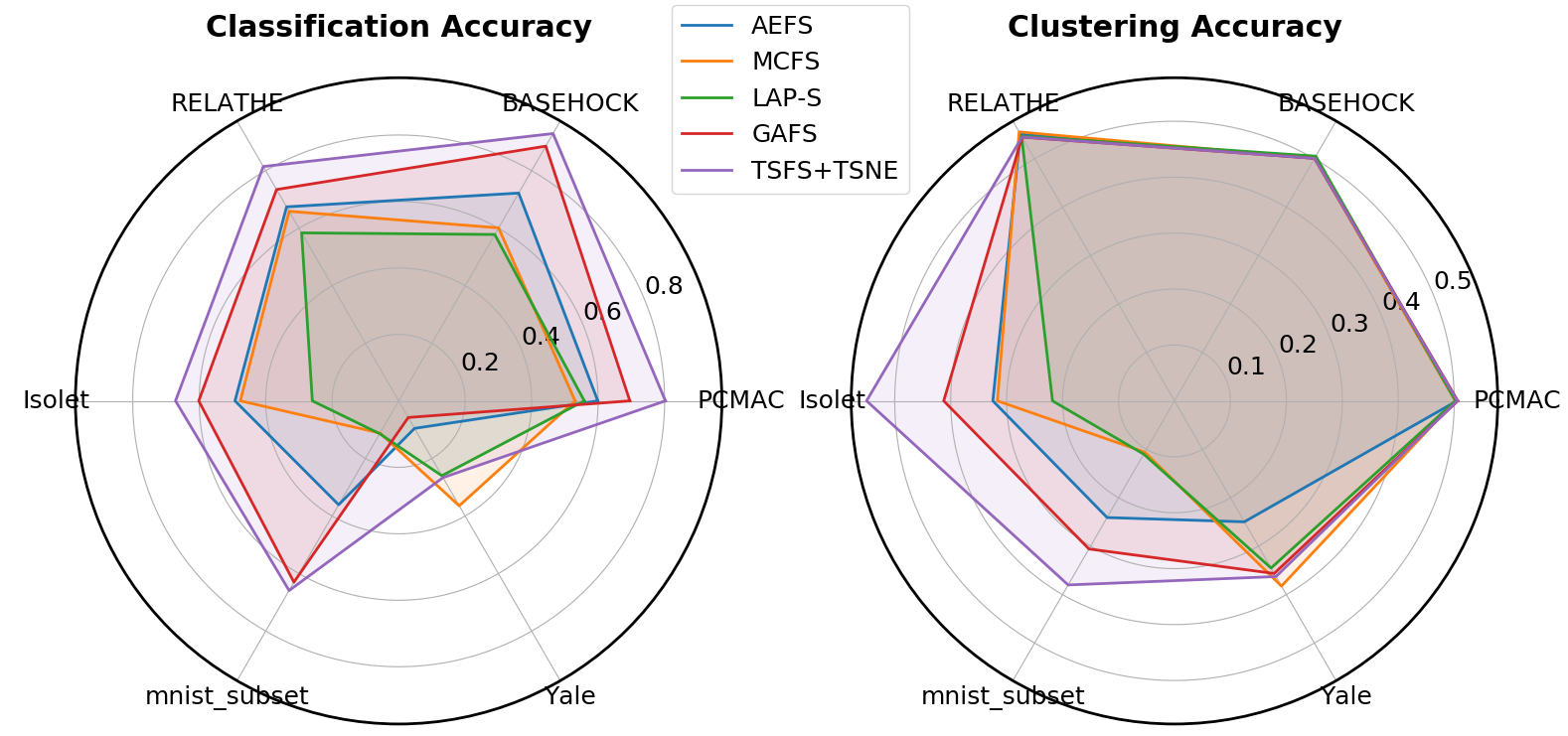}
    \caption{The comparison of methods for $2\%$ of features in terms of classification and clustering accuracy. When the number of selected features are few the correct selection is so important.}
    \label{fig:spyder}
\end{figure}

As describe in Section \ref{method}, we propose a general pipeline for TSFS. In this pipeline, the teacher is not specified because in various cases or datasets it can be different. To perform a fair comparison of our method with other existing method, one has to fix the teacher method. To this end, first we evaluate different teacher methods to find the optimal one for selected datasets. Then the performance of TSFS is compared to the results of other existing approaches.

For \emph{student network} (feature selection network), the network structure and parameters are defined as:
\begin{enumerate}
    \item Single hidden layer
    \item The number neurons in hidden layer is $20$
    \item The Adam optimizer is used for optimization and training.
    \item The epoch and batch size is considered as 500 and 32, respectively.
    \item The regularization parameter is $\lambda = 0.1$
\end{enumerate}

\textbf{First Step: Evaluation of Different Teacher Methods}: 

In this part, \textcolor{black}{to select a good teacher method,} we evaluate our method using 5 well-known manifold leaning schemes of MDS \cite{msd}, Isomap \cite{isomap}, TSNE \cite{tsne}, Spectral Embedding(SE) \cite{se} and Local Linear Embedding(LLE) \cite{lle}. The subspace (embedding) dimension for all methods is experimentally chosen as 2. 

Figure \ref{fig:my_results} depicts the performance (in terms of classification accuracy) of TSFS using different manifold learning methods as the teacher. As shown, at least for these datasets, the TSNE performs better than others. \textcolor{black}{So in the next part, we only use the TSNE manifold learning as teacher to compare the performance of TSFS with other} unsupervised feature selection methods.

Figure \ref{fig:reconst} shows the visual reconstruction results on 6 image of COIL20 dataset. As evident, the proposed algorithm with TSNE as teacher successfully selects the important features so that the original image can be well reconstructed. It is clear that the method selects most of the edge features that are necessary for recognition and reconstruction of the object. \textcolor{black}{It worth mentioning that selection of TSFS is not mandatory for the proposed scheme and considering the application, if any other scheme is more suitable it can be used as the \emph{teacher} method.}

\textbf{Second Step: Comparison with Other Existing Methods}:

To show the performance of our method against other existing methods, we use TSNE as the teacher method and compare \textcolor{black}{the performance metrics explained in Section \ref{section:metrics} for TSFS and five recent} unsupervised feature selection methods: Laplacian Score, AEFS, MCFS and GAFS. These methods are described in Section \ref{section:related}. 

Figures \ref{fig:res_classification}, \ref{fig:res_clustering}, \ref{fig:res_NMI} and \ref{fig:res_mse} shows the classification, clustering and reconstruction results. As depicted, the proposed TSFS method in most cases outperform other existing methods. 

\textcolor{black}{To better compare the results, especially in the region where only very few features are selected (small $p$) } we additionally compare  all methods when the percentage of selected features is two. The results are shown in Figure \ref{fig:spyder} as a spider chart where the occupied aria of our method is larger than other approaches. Furthermore, in all datasets , except for the Yale dataset and for the MCFS method, the proposed method has a better performance compared with others.

At the end, for illustration, Figure \ref{fig:mnist} shows the reconstruction result using limited selected features for different unsupervised feature selection methods. The reconstruction network is the same for all methods as described in Section \ref{section:metrics}. As can be seen, the TSFS+TSNE method successfully reconstructs all digits (except for 5). The performance superiority of the method is more evident in digits 1,3,6 and 9.

\subsubsection{Supervised Teacher-Student Feature Selection(S-TSFS)}

\begin{figure}
    \centering
    \includegraphics[width=1\textwidth]{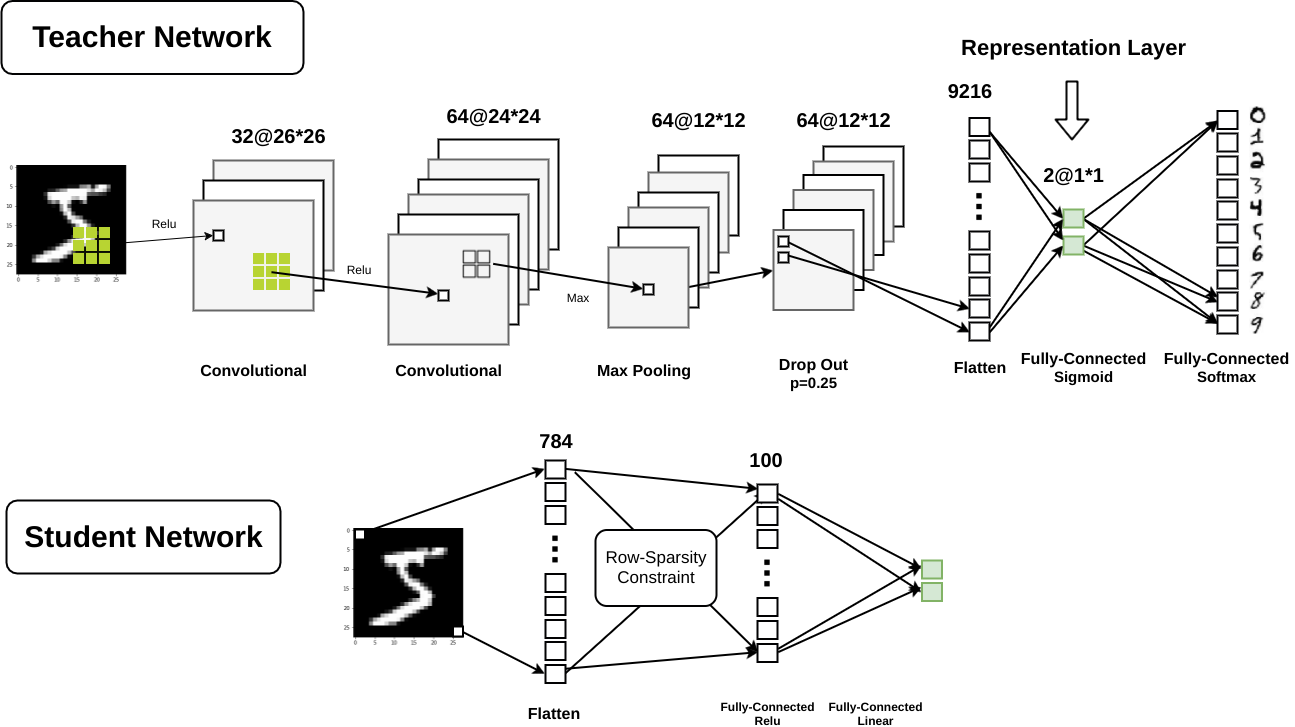}
    \caption{The CNN architecture used as the teacher in mnist for supervised feature selection}
    \label{fig:arch}
\end{figure}

\textcolor{black}{If the teacher is a supervised method then the feature selection will be a supervised feature selection approach}. In this part we use a convolutional neural network (CNN) to train a suitable representation from MNIST dataset in a supervised manner using labels. The structure of the employed deep neural network is shown in Figure \ref{fig:arch}. After training this network on all 60,000 images of the MNIST dataset, a subset of dataset (1000 samples) are fed into the network, and using the obtained representation, the feature selection network is trained as describe in Section \ref{method}. 

\begin{figure}
    \centering
    \includegraphics[width=1\textwidth]{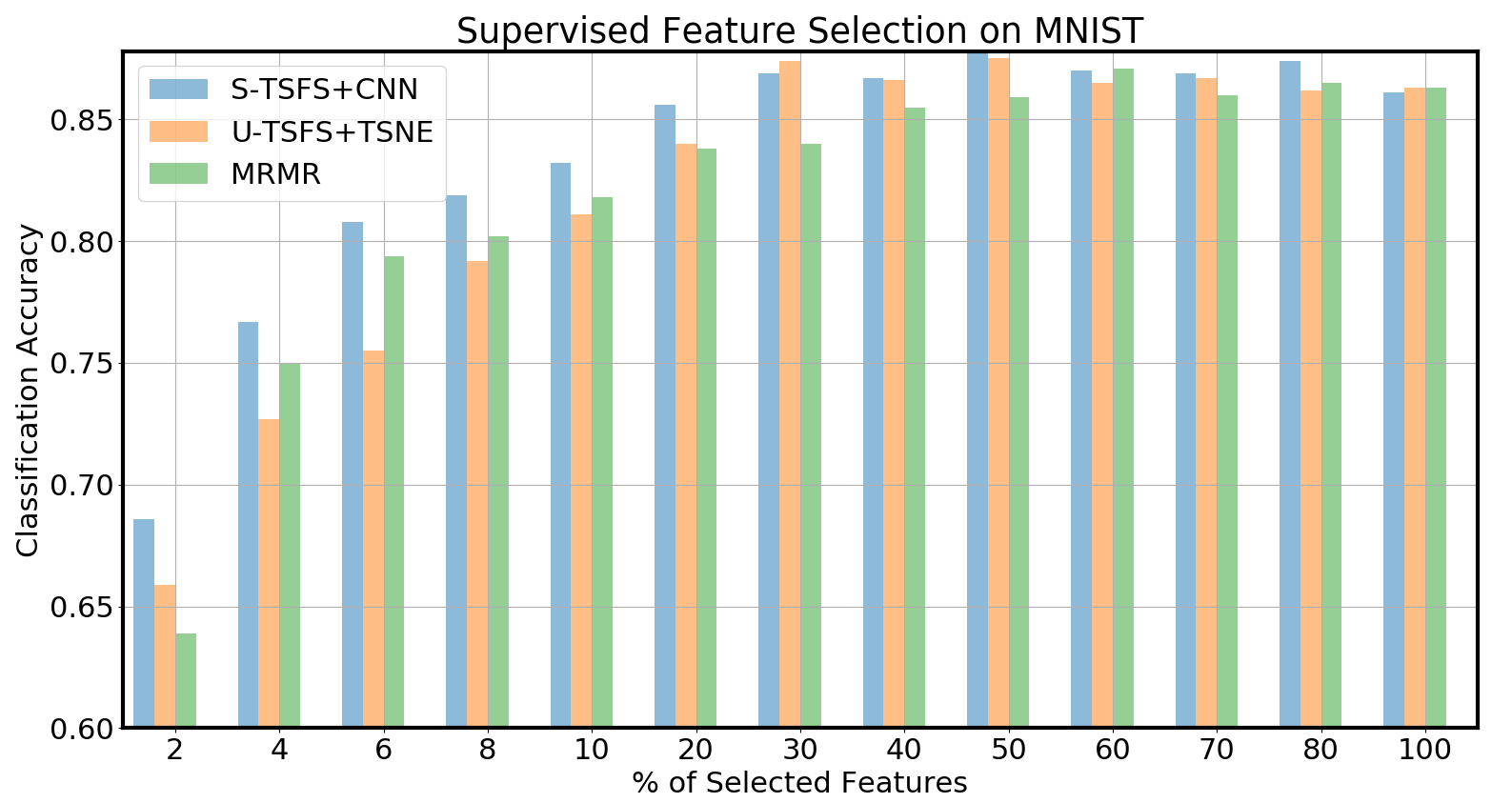}
    \caption{The classification accuracy on mnist\_subset dataset}
    \label{fig:supervised}
\end{figure}

For a fair performance comparison, this approach should be compared with other supervised feature selection methods. As described in Section \ref{section:related}, the \textcolor{black}{information-based} methods are usually supervised. Figure \ref{fig:supervised} shows the classification accuracies for the U-TSFS, the S-TSFS and the MRMR using the MLP classifier described in Section \ref{section:metrics}. Evidently, in most cases, the S-TSFS has a better performance compared to other ones. It is worth mentioning that $50\%$ selection of features have the best performance because in the MNIST dataset almost $50\%$ of features are black (for all samples). Thus these features are redundant and including them in the classification stage can increase the complexity of the classifier and so decrease its generalization ability which leads to worse performances.

\subsection{Sensitivity Analysis}

\begin{figure}
    \centering
    \includegraphics[width=1\textwidth]{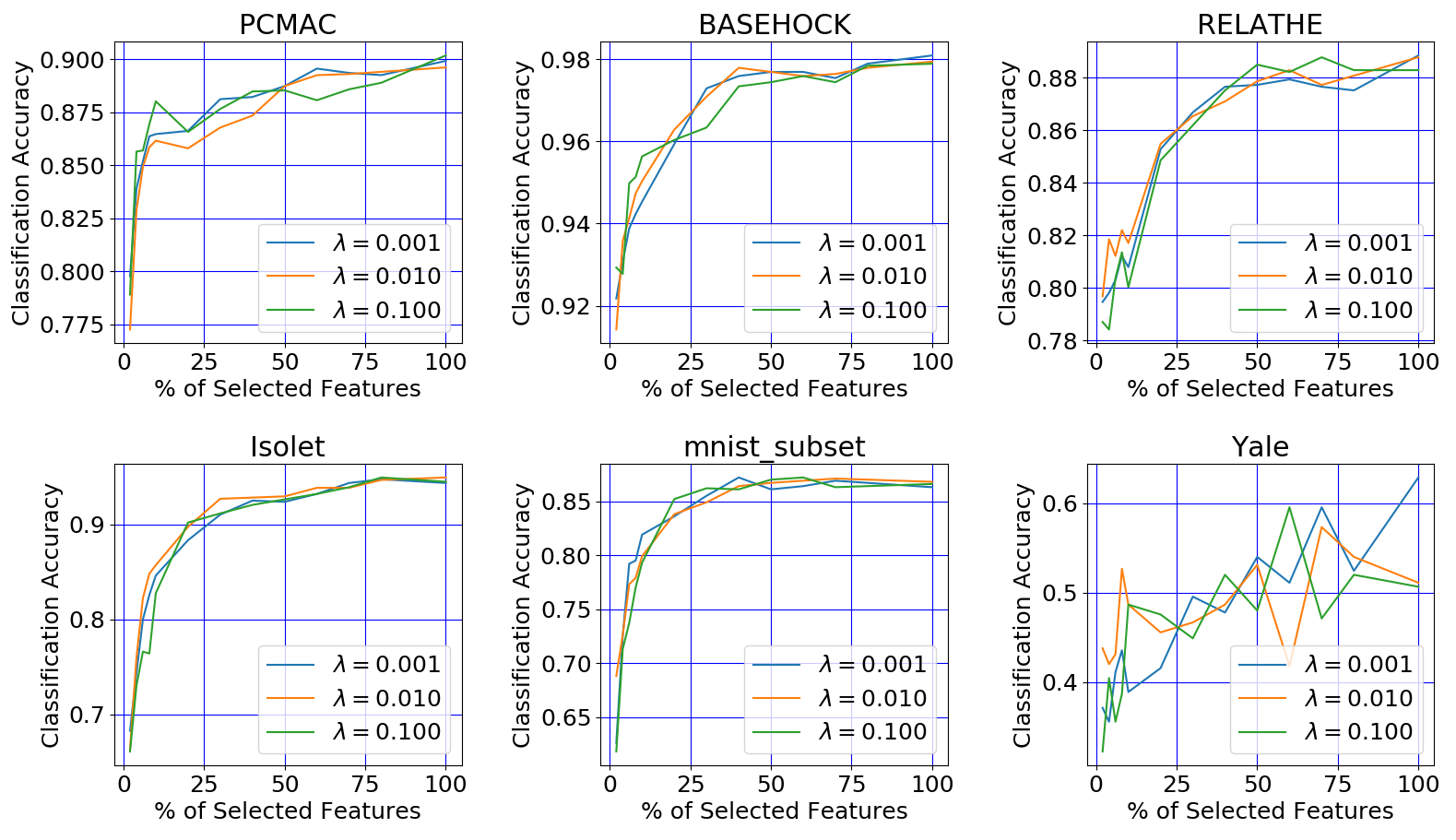}
    \caption{The sensitivity analysis of $\lambda$ parameter for TSFS+TSNE  }
    \label{fig:sensitivity}
\end{figure}

In the proposed algorithm, there are two parameters to be adjusted. The first one is the number of neurons in the hidden layer of the feature selection network and the second one is the trade-off parameter ($\lambda$). The number of hidden neurons should be set based on the representation dimension. Experimentally, we developed a rule of thumb for its value which is ten times the representation dimension. In our experiments, the representation dimension is two so the number of hidden neurons is selected as 20. 

The other parameter is $\lambda$ for which we perform a sensitivity analysis. Figure \ref{fig:sensitivity} shows the classification accuracies for different datasets with three different values for the $\lambda \in \{ 0.001,0.01,0.1 \} $  parameter. As can be seen for all 3 values of $\lambda$, the results are not much different \textcolor{black}{which verifies that the algorithm is not very sensitive to the selected value of $\lambda$}. As mentioned in Section \ref{section:unsupervised} we use $\lambda=0.1$ for all experiments.

\section{Conclusion}
\label{section:conclusion}
In this paper, we presented a teacher-student scheme for deep feature selection (TSFS). \textcolor{black}{The idea is to partition the feature selection task into, first, finding a good low-order representation and then use this representation for feature selection}. The proposed approach can be applied to both supervised and unsupervised applications. The extensive experiments, performed on different datasets, verifies the superiority of the TSFS against other existing methods in terms of classification accuracy, clustering performance (accuracy and NMI) and reconstruction performance (MSE) using limited selected features. 
\section*{References}

\bibliography{ref}

\begin{thebibliography}{10}
\expandafter\ifx\csname url\endcsname\relax
  \def\url#1{\texttt{#1}}\fi
\expandafter\ifx\csname urlprefix\endcsname\relax\def\urlprefix{URL }\fi
\expandafter\ifx\csname href\endcsname\relax
  \def\href#1#2{#2} \def\path#1{#1}\fi

\bibitem{zhang2018self}
R.~Zhang, F.~Nie, X.~Li, Self-weighted supervised discriminative feature
  selection, IEEE transactions on neural networks and learning systems 29~(8)
  (2018) 3913--3918.

\bibitem{gan2018supervised}
J.~Gan, G.~Wen, H.~Yu, W.~Zheng, C.~Lei, Supervised feature selection by
  self-paced learning regression, Pattern Recognition Letters.

\bibitem{sechidis2018simple}
K.~Sechidis, G.~Brown, Simple strategies for semi-supervised feature selection,
  Machine Learning 107~(2) (2018) 357--395.

\bibitem{sheikhpour2017survey}
R.~Sheikhpour, M.~A. Sarram, S.~Gharaghani, M.~A.~Z. Chahooki, A survey on
  semi-supervised feature selection methods, Pattern Recognition 64 (2017)
  141--158.

\bibitem{panday2018feature}
D.~Panday, R.~C. de~Amorim, P.~Lane, Feature weighting as a tool for
  unsupervised feature selection, Information Processing Letters 129 (2018)
  44--52.

\bibitem{li2018unsupervised}
Y.~Li, C.~Lei, Y.~Fang, R.~Hu, Y.~Li, S.~Zhang, Unsupervised feature selection
  by combining subspace learning with feature self-representation, Pattern
  Recognition Letters 109 (2018) 35--43.

\bibitem{qi2018unsupervised}
M.~Qi, T.~Wang, F.~Liu, B.~Zhang, J.~Wang, Y.~Yi, Unsupervised feature
  selection by regularized matrix factorization, Neurocomputing 273 (2018)
  593--610.

\bibitem{zhu2018co}
P.~Zhu, Q.~Xu, Q.~Hu, C.~Zhang, Co-regularized unsupervised feature selection,
  Neurocomputing 275 (2018) 2855--2863.

\bibitem{mirzaei2017variational}
A.~Mirzaei, Y.~Mohsenzadeh, H.~Sheikhzadeh, Variational relevant sample-feature
  machine: a fully bayesian approach for embedded feature selection,
  Neurocomputing 241 (2017) 181--190.

\bibitem{mohsenzadeh2016incremental}
Y.~Mohsenzadeh, H.~Sheikhzadeh, S.~Nazari, Incremental relevance sample-feature
  machine: A fast marginal likelihood maximization approach for joint feature
  selection and classification, Pattern Recognition 60 (2016) 835--848.

\bibitem{han2018autoencoder}
K.~Han, Y.~Wang, C.~Zhang, C.~Li, C.~Xu, Autoencoder inspired unsupervised
  feature selection, in: 2018 IEEE International Conference on Acoustics,
  Speech and Signal Processing (ICASSP), IEEE, 2018, pp. 2941--2945.

\bibitem{feng2018graph}
S.~Feng, M.~F. Duarte, Graph autoencoder-based unsupervised feature selection
  with broad and local data structure preservation, Neurocomputing.

\bibitem{chang2017dropout}
C.-H. Chang, L.~Rampasek, A.~Goldenberg, Dropout feature ranking for deep
  learning models, arXiv preprint arXiv:1712.08645.

\bibitem{hinton2015distilling}
G.~Hinton, O.~Vinyals, J.~Dean, Distilling the knowledge in a neural network,
  arXiv preprint arXiv:1503.02531.

\bibitem{he2006laplacian}
X.~He, D.~Cai, P.~Niyogi, Laplacian score for feature selection, in: Advances
  in neural information processing systems, 2006, pp. 507--514.

\bibitem{peng2005feature}
H.~Peng, F.~Long, C.~Ding, Feature selection based on mutual information
  criteria of max-dependency, max-relevance, and min-redundancy, IEEE
  Transactions on pattern analysis and machine intelligence 27~(8) (2005)
  1226--1238.

\bibitem{cai2010unsupervised}
D.~Cai, C.~Zhang, X.~He, Unsupervised feature selection for multi-cluster data,
  in: Proceedings of the 16th ACM SIGKDD international conference on Knowledge
  discovery and data mining, ACM, 2010, pp. 333--342.

\bibitem{zhu2015unsupervised}
P.~Zhu, W.~Zuo, L.~Zhang, Q.~Hu, S.~C. Shiu, Unsupervised feature selection by
  regularized self-representation, Pattern Recognition 48~(2) (2015) 438--446.

\bibitem{radford2015unsupervised}
A.~Radford, L.~Metz, S.~Chintala, Unsupervised representation learning with
  deep convolutional generative adversarial networks, arXiv preprint
  arXiv:1511.06434.

\bibitem{kingma2014adam}
D.~P. Kingma, J.~Ba, Adam: A method for stochastic optimization, arXiv preprint
  arXiv:1412.6980.

\bibitem{kuhn1955hungarian}
H.~W. Kuhn, The hungarian method for the assignment problem, Naval research
  logistics quarterly 2~(1-2) (1955) 83--97.

\bibitem{msd}
F.~W. Young, Multidimensional scaling: History, theory, and applications,
  Psychology Press, 2013.

\bibitem{isomap}
J.~B. Tenenbaum, V.~De~Silva, J.~C. Langford, A global geometric framework for
  nonlinear dimensionality reduction, science 290~(5500) (2000) 2319--2323.

\bibitem{tsne}
L.~v.~d. Maaten, G.~Hinton, Visualizing data using t-sne, Journal of machine
  learning research 9~(Nov) (2008) 2579--2605.

\bibitem{se}
M.~Belkin, P.~Niyogi, Laplacian eigenmaps and spectral techniques for embedding
  and clustering, in: Advances in neural information processing systems, 2002,
  pp. 585--591.

\bibitem{lle}
S.~T. Roweis, L.~K. Saul, Nonlinear dimensionality reduction by locally linear
  embedding, science 290~(5500) (2000) 2323--2326.

\end{thebibliography}

\end{document}